\documentclass{article} \usepackage{iclr2027_conference,times}

\usepackage{amsmath,amsfonts,bm}

\def\eqref#1{equation~\ref{#1}}

\def\1{\bm{1}}

\DeclareMathAlphabet{\mathsfit}{\encodingdefault}{\sfdefault}{m}{sl}
\SetMathAlphabet{\mathsfit}{bold}{\encodingdefault}{\sfdefault}{bx}{n}

\usepackage{hyperref}
\usepackage{url}

\title{When KL Regularization Misfires in Group Policy Optimization}

\author{Fei Ding\thanks{Corresponding author: \texttt{dignfei@gmail.com}.} \\
Alibaba Group
}

\newcommand{\method}{ZCPO}
\usepackage{booktabs}
\usepackage{threeparttable}
\usepackage{tabularx}
\usepackage{longtable}

\iclrfinalcopy \usepackage{graphicx}
\begin{document}

\maketitle
\lhead{Preprint}

\begin{abstract}
Why does removing reference-policy KL regularization sometimes improve group policy optimization~\citep{ICLR2026_82846e19}? This motivates studying how reference-policy information should enter group-relative updates. We analyze seven potential failure modes in KL--reward interactions: residual KL updates after reward clipping, after gradient cancellation, and in groups with identical rewards; KL growth with response length and an imbalance in its relative contribution; KL concentration on a small number of tokens; and sampling noise when k1 is incorporated into rewards. We propose Zero-Sum Calibrated Policy Optimization (ZCPO), which uses relative drift measured by conditional KL to calibrate within-group reward coefficients and integrates them into the base surrogate. Mathematical reasoning experiments and ablations support this design's effectiveness in our settings.
\end{abstract}

\begin{figure}[!ht]
\centering
\includegraphics[width=\textwidth]{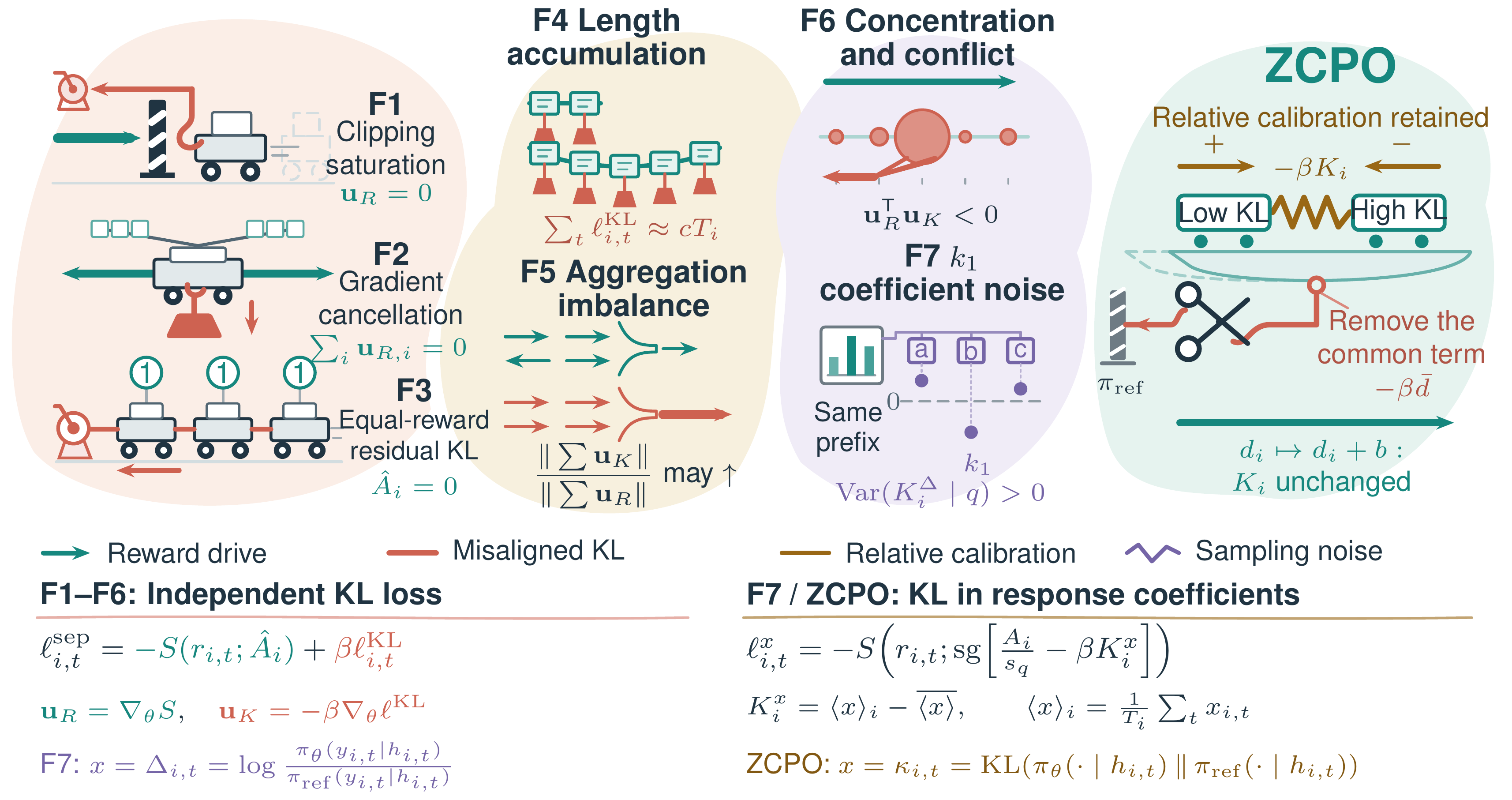}
\caption[]{\textbf{KL placement and seven types of mismatch.}
The right panel illustrates the coefficient decomposition ($d_i=\langle\kappa\rangle_i=\widetilde D_i$, $K_i^\kappa=K_i$): a uniform shift represents the common KL offset, and the gold arrows indicate the relative calibration contribution.
$S$ denotes the base surrogate, $\operatorname{sg}$ stops gradients, $r=\pi_\theta/\pi_{\rm old}$ is the probability ratio for a sampled token, and $\beta\ge0$ is the KL weight; the independent term is illustrated using $\ell^{\rm KL}=e^{-\Delta}+\Delta-1$ ($k_3$).
$A_i=R_i-\bar R$, $\hat A_i=A_i/(\sigma_q+\varepsilon_{\rm adv})$, and $s_q=\max_j|A_j|$; $R,\sigma_q,T,h,G,q$ denote the reward, reward standard deviation, length, prefix, group size, and prompt, respectively; overbars indicate group means, and $\varepsilon_{\rm adv}>0$.
Conditions illustrated: F2 assumes the same context/token, all gates open, and equal weights; F3 assumes unfiltered groups; F4 assumes no length averaging and $\ell^{\rm KL}_{i,t}\approx c>0$; F5 assumes nonzero denominators; F7 assumes $G\ge2$ independent single-token responses sampled from the current policy for the same prompt, with distinct, full-support $\pi_\theta,\pi_{\rm ref}$. The coefficient objective first skips groups with $G<2$ or $s_q=0$.}
\label{fig:zcpo-overview}
\end{figure}

\section{Introduction}

Group-relative policy optimization trains language models by comparing rewards within a group, eliminating the need for a value model~\citep{shao2024deepseekmathpushinglimitsmathematical}. Recent methods improve training through clipping, normalization, and gradient aggregation; among them, DAPO and GMPO omit an explicit reference KL penalty~\citep{NEURIPS2025_a4277440,ICLR2026_44a1f18a}. In controlled experiments, PAPO removes the reference KL from GRPO, increasing the overall average multimodal reasoning accuracy (avg@8) of Qwen2.5-VL-3B and 7B from $47.92\%$ to $50.18\%$ and from $58.78\%$ to $61.30\%$, respectively~\citep[Appendix F, Table 4]{ICLR2026_82846e19}. Text reasoning ablations in Open-Reasoner-Zero also report that PPO without KL outperforms variants that add KL to the loss or reward under their evaluation settings~\citep{hu2025openreasonerzeroopensourceapproach}.

The observation that removing reference KL can help training raises a broader design question: how should reference-policy information enter group-relative optimization? We analyze the interaction between independent KL and reward updates, and study within-group relative calibration using reference-policy drift. Specifically, our contributions are as follows:
\begin{itemize}
 \item We consolidate seven potential failure modes of KL in GRPO and its variants. These modes characterize residual regularization after reward gradients vanish, KL imbalances across lengths and positions, and sampling noise in KL estimation.
 \item We propose \method{}, which uses conditional KL to calibrate group-relative reward coefficients, and characterize coefficient optimality for fixed inputs; mathematical reasoning experiments and ablations support this design's effectiveness in our settings.
\end{itemize}

\section{Related Work}
\label{sec:related}

\textbf{Variants of group-relative policy optimization.} With a fixed reference, GRPO adds a per-token k3 KL penalty relative to the reference model to the loss~\citep{shao2024deepseekmathpushinglimitsmathematical}. Subsequent work primarily modifies clipping, normalization, and aggregation: DAPO introduces clip-higher, dynamic sampling, and token-level aggregation~\citep{NEURIPS2025_a4277440}; GSPO uses sequence-level importance ratios~\citep{zheng2025groupsequencepolicyoptimization}; GMPO uses geometric-mean aggregation~\citep{ICLR2026_44a1f18a}; and a range of methods improve clipping or trust regions~\citep{ICLR2026_cca4df59,ICLR2026_124dde49,lee2026quatroqueryadaptivetrustregion,wang2026anchoredpolicyoptimizationmitigating,cai2026euclideanclippingovercomingexploration}. These studies improve group policy optimization from different perspectives; our focus is the interaction between the KL term and group-relative reward updates and the points at which it may fail. Adaptive AGPO uses statistics such as KL drift to adjust clipping and sampling temperature~\citep{hu2026agpoadaptivegrouppolicy}; asymmetric AGPO constrains group-relative advantages and strengthens negative updates for incorrect responses~\citep{xu2026agpoasymmetricgrouppolicy}. For groups with limited information, existing methods recover learning signals through sampling allocation, trajectory filtering, or finer-grained rewards~\citep{sudalairaj2026sgpotradinginferenceflops,lin2025cppoacceleratingtraininggroup,deng2026prismgrpofastervlapolicy}. \citet{bae-etal-2026-online} retain training prompts with intermediate success rates through online difficulty filtering. GOPO replaces explicit KL regularization with a quadratic probability-ratio penalty~\citep{zixian2026grouporthogonalizedpolicyoptimizationgroup}.

\textbf{KL estimators and placement.} \citet{vojnovic2025alignmentobjectivegrpo} analyze how reward normalization and reference-policy penalties jointly determine GRPO's stationary policy. \citet{tang2025pitfallskldivergencegradient} analyze bias in KL gradient estimation and discuss leave-one-out baselines and causal accumulation, but focus primarily on recovering the gradient of a given KL objective rather than the interaction of independent KL with reward clipping or group filtering. \citet{shah2026comedyestimatorsklregularization} compare the gradient bias of k1/k3 in rewards or losses under current-policy sampling and evaluate performance in asynchronous training, but do not discuss constructing conditional KL as a within-group zero-sum correction coefficient. The online algorithms of GVPO and RPG use an old policy that is updated throughout training as the reference~\citep{NEURIPS2025_f21a76d6,ICLR2026_8e49d32f}; from the perspective of F6 in this paper, this may alleviate the persistent constraint that a fixed initial reference imposes on policy changes already learned. However, for constructions based on sampled log-ratios, refreshing the reference does not guarantee the elimination of subsequent conditional sampling fluctuations, nor does within-group centering guarantee their elimination.

\textbf{Conditional KL and local constraint adjustment.} \citet{NEURIPS2025_a2e9be3d} use per-prefix conditional KL to reduce the variance of numerical estimation and full-sequence KL gradient estimation, but do not study a joint coefficient design combining length normalization, within-group centering, and gating on reward-active groups. \citet{vassoyan-etal-2025-ignore} weight per-token KL by the reference policy's entropy to relax constraints at highly uncertain positions, but this confidence-based rule does not address cases in which the reference policy is overconfident.

\section{Seven Potential Failure Modes in the Interaction Between KL and Rewards in GRPO and Its Variants}
\label{sec:mismatches}

This section characterizes differences in how KL signals and group-relative reward updates act under specified conditions, and uses these differences to formulate mechanistic hypotheses about their effects on training. We first analyze the case where per-token KL is added as an independent loss term, then discuss variants that incorporate the k1 sampled log-ratio into rewards in F7; the conditions for each mode are stated separately. GRPO uses an independent KL loss~\citep{shao2024deepseekmathpushinglimitsmathematical}. The subsequent experiments report structural diagnostics, F1 mechanism controls, and the task performance of the full method. Appendix~\ref{app:cross-model-task} reports the F3 group-gating ablation.

\paragraph{Problem setup and notation.}
For the same prompt $q$, the sampling policy $\pi_{\rm old}$ generates $G\ge2$ responses $\{y_i\}_{i=1}^G$. Response $i$ has length $T_i$, token $y_{i,t}$ at position $t$, context $h_{i,t}=(q,y_{i,<t})$, and reward $R_i$. The current policy $\pi_\theta$ is parameterized by $\theta$, and $\pi_{\rm ref}$ is the reference policy; within a single update, the sampling policy, reference policy, samples, and rewards are all fixed. Let the within-group reward mean be $\bar R=G^{-1}\sum_iR_i$ and the standard deviation be $\sigma_q=\operatorname{std}(R_1,\ldots,R_G)$. The baseline's group-normalized advantage and token probability ratio are
$$
\hat A_i=\frac{R_i-\bar R}{\sigma_q+\varepsilon_{\rm adv}},\qquad
r_{i,t}=\frac{\pi_\theta(y_{i,t}\mid h_{i,t})}{\pi_{\rm old}(y_{i,t}\mid h_{i,t})},
$$
where $\varepsilon_{\rm adv}>0$ is a numerical stability constant; groups with identical rewards satisfy $\hat A_i=0$, and every group satisfies $\sum_i\hat A_i=0$. No gradient is taken through $\hat A_i$ when computing the policy gradient. We further define the token score-function gradient and the sampled log-ratio relative to the reference policy as
$$
g_{i,t}=\nabla_\theta\log\pi_\theta(y_{i,t}\mid h_{i,t}),\qquad
\Delta_{i,t}=\log\frac{\pi_\theta(y_{i,t}\mid h_{i,t})}{\pi_{\rm ref}(y_{i,t}\mid h_{i,t})}.
$$

\paragraph{Baseline with an independent KL loss.}
Let $\beta\ge0$ be the KL regularization coefficient. For a base algorithm using per-token probability ratios, let $S_{\rm base}(r;a)$ denote its existing policy optimization surrogate objective, where $a$ is a fixed advantage coefficient. Whether clipping is used, and the specific clipping rule, are determined by the base algorithm; ZCPO supplies a KL correction coefficient to this surrogate. Let $\ell^{\rm KL}_{i,t}$ be a differentiable per-token KL regularization loss that is independent of the reward coefficient and its clipping gate. The independently additive baseline objective is
$$
\ell_{i,t}=-S_{\rm base}(r_{i,t};\hat A_i)
+\beta\,\ell^{\rm KL}_{i,t}.
$$

\textbf{Lemma (F1--F3: residual KL after reward gradients vanish).} Using the notation above, consider a piecewise surrogate satisfying $\partial_r S_{\rm base}(r;a)=a\,m_{\rm base}(r,a)$, where $m_{\rm base}(r,a)\in\{0,1\}$ denotes the base algorithm's gradient gate: it equals 1 on a linear branch and 0 on a branch that is constant with respect to $r$. By the chain rule, at differentiable points within each piece,
\begin{equation}
\label{eq:residual-gradient}
\nabla_\theta\ell_{i,t}=
\underbrace{-\hat A_i\,r_{i,t}\,g_{i,t}\,m_{\rm base}(r_{i,t},\hat A_i)}_{\text{reward term}}
+\underbrace{\beta\nabla_\theta\ell^{\rm KL}_{i,t}}_{\text{KL term}}.
\end{equation}
When $\beta>0$ and $\nabla_\theta\ell^{\rm KL}_{i,t}\neq0$, the KL term is nonzero. Thus, an independent KL gradient may remain after the reward gradient vanishes, including in the following three cases.

\textbf{F1: residual KL-driven updates after reward clipping.} When clipping by the base algorithm makes the reward surrogate locally flat, $m_{\rm base}(r_{i,t},\hat A_i)=0$: the reward term is zero while the KL term remains unchanged, so the token's direct gradient comes entirely from the independent KL branch.

\textbf{F2: residual KL after reward gradient cancellation.} When within-group reward gradients cancel during aggregation, independent KL gradients need not cancel simultaneously, so KL-driven updates may remain. With all gradient gates open (the reward branch is unclipped) and equal-weight aggregation, reward gradients at prefix positions shared by the entire group, with identical sampled tokens, cancel exactly because the advantages sum to zero, providing an exact special case. Whether identical segments in different contexts yield approximate cancellation depends on their actual gradients and reward coefficients.

\textbf{F3: residual KL-driven updates in groups with identical rewards.} When all $R_i$ in a group are equal and the group still participates in the update, $\hat A_i\equiv0$, so the reward term is zero for every token. The group's direct gradient contains only the independent KL term and may remain nonzero. $\square$

\paragraph{F4: KL modulates language model response length.}
Let the conditional KL at each position be
$$
\kappa_{i,t}=\mathrm{KL}\big(\pi_\theta(\cdot\mid h_{i,t})\|\pi_{\rm ref}(\cdot\mid h_{i,t})\big).
$$
Since $\kappa_{i,t}\ge0$, sequence KL accumulates along a response; without length normalization, longer responses incur larger KL penalties when per-token KL values are comparable and positive. An independent KL loss can also affect response length through the gradient at the stopping position in Eq.~(\ref{eq:residual-gradient}).

\paragraph{F5: the relative contribution of KL may increase as more tokens are aggregated.}
The two gradient branches in Eq.~(\ref{eq:residual-gradient}) may accumulate at different rates because they undergo different degrees of cancellation. If reward gradients grow more slowly due to cancellation while KL gradients accumulate more rapidly, the gradient contribution of KL relative to that of rewards may increase as more tokens participate in aggregation.

\paragraph{F6: potential suppression of beneficial policy changes when KL signals are concentrated.}
If KL signals are concentrated on a small number of tokens, their local pull toward the reference may weaken reward update signals at those positions when it conflicts with reward updates. Taking the per-token k3 KL loss as an example,
$$
\ell^{\rm KL}_{i,t}=e^{-\Delta_{i,t}}+\Delta_{i,t}-1,
\qquad
\beta\nabla_\theta\ell^{\rm KL}_{i,t}=\beta(1-e^{-\Delta_{i,t}})g_{i,t}.
$$
The nonzero local descent direction of this term pulls the sampled token's probability toward the reference policy. When the reward update pushes it further away from the reference, the two directions oppose each other, potentially suppressing beneficial policy changes. Comparisons of KL signal concentration and drift are presented in Section~\ref{sec:exp}.

\paragraph{F7: conditional sampling noise in the k1 coefficient.}
Even when all responses have the same length-normalized conditional KL, response-level KL coefficients constructed using k1 may still contain noise introduced by token sampling, and within-group centering does not guarantee its removal. For example, consider $G\ge2$ independent single-token responses sampled from the current policy for the same prompt $q$. If $\pi_\theta(\cdot\mid q)$ and $\pi_{\rm ref}(\cdot\mid q)$ have full support and differ, then
$$
\operatorname{Var}\!\Big(\Delta_{i,1}-\frac1G\sum_j\Delta_{j,1}\Big)
=\Big(1-\frac1G\Big)\operatorname{Var}(\Delta_{i,1})>0 .
$$
Here, $\kappa_{i,1}$ is identical for all $i$ (the context is the same), and all differences arise from sampling.

\section{Zero-Sum Calibrated Policy Optimization (ZCPO)}
\label{sec:method}

\begin{figure}[!ht]
\centering
\includegraphics[width=\textwidth]{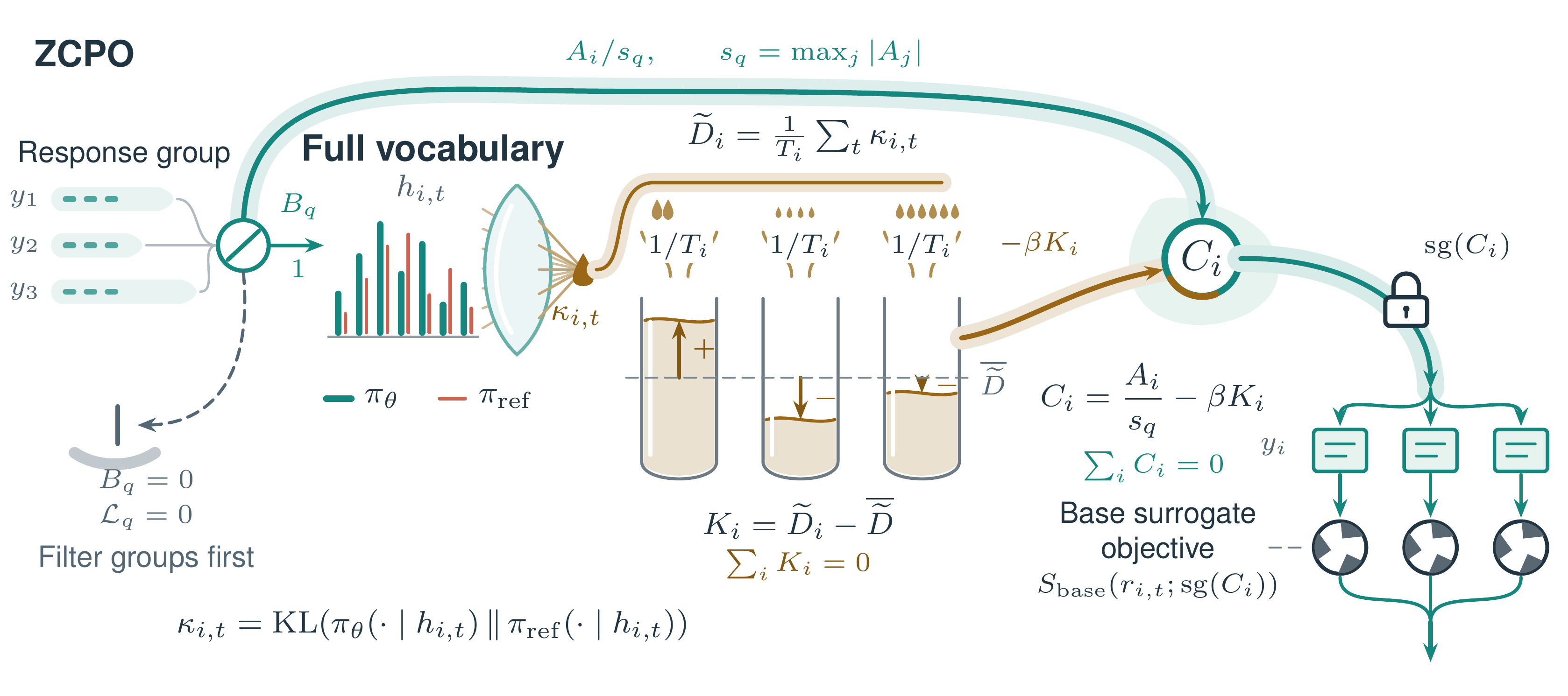}
\caption[]{\textbf{ZCPO coefficient construction.} $A_i=R_i-\bar R$, $B_q=\mathbf1\{G\ge2,\ s_q>0\}$. All tokens in a response share $\operatorname{sg}(C_i)$. Zero-sum coefficients do not imply a zero total gradient; conditional KL eliminates only token sampling noise conditional on a given prefix.}
\label{fig:zcpo-method}
\end{figure}

ZCPO uses conditional KL as reference information to calibrate response-level update coefficients relative to their group, then incorporates them into the base algorithm (Figure~\ref{fig:zcpo-method}).

We first compute
$$
A_i=R_i-\bar R,\qquad s_q=\max_i|A_i|,\qquad
B_q=\mathbf1\{G\ge2\}\mathbf1\{s_q>0\}.
$$
If $B_q=0$, we directly return zero loss without computing $1/s_q$. The following quantities are computed only for groups with $B_q=1$:
$$
\begin{gathered}
D_i=\sum_{t=1}^{T_i}\kappa_{i,t},\qquad
\widetilde D_i=\frac{D_i}{T_i},\qquad
K_i=\widetilde D_i-\frac1G\sum_j\widetilde D_j,\\
\boxed{C_i=\frac{A_i}{s_q}-\beta K_i}.
\end{gathered}
$$
Here, $\kappa_{i,t}$ is computed by summing over the full vocabulary distributions of the current and reference policies; $1/s_q$ scales only the reward term. Conditional KL can reuse the forward-pass outputs of the current and reference policies, with additional computation consisting mainly of vocabulary reductions; implementation details are provided in Appendix~\ref{app:exp-config}. The coefficient $A_i/s_q$ is normalized by the maximum absolute centered reward, unlike the baseline coefficient $\hat A_i$, which uses standard-deviation normalization.

\paragraph{Integration with the base algorithm.}
After computing $C_i$ using the current policy, we hold it fixed during the current backward pass and use it to replace the reward coefficient in the base surrogate. For active groups,
$$
\mathcal L_q=-\sum_{i=1}^G\sum_{t=1}^{T_i}
w_{i,t}\,S_{\rm base}\big(r_{i,t};\operatorname{sg}(C_i)\big),
$$
where $\operatorname{sg}$ denotes stop-gradient and $w_{i,t}\ge0$ are aggregation weights held fixed during the current update. This expression defines the policy surrogate objective with frozen coefficients: gradients propagate only through $r_{i,t}$ and exclude $\nabla_\theta C_i$. The surrogate objective is updated whenever the coefficients are recomputed. The base algorithm supplies the probability-ratio treatment, clipping, and aggregation rules.

\paragraph{Mechanistic hypothesis and coefficient calibration.}
We construct the coefficients under the following mechanistic hypothesis: keep update coefficients close to the normalized reward coefficients while maintaining a within-group zero sum and favoring responses with lower mean KL. Fix $A_i/s_q$ and $\widetilde D_i$ in an active group, let $\boldsymbol c=(c_1,\ldots,c_G)\in\mathbb R^G$ denote the update coefficients to be calibrated, and consider
\begin{equation}
\label{eq:coefficient-calibration}
\min_{\boldsymbol c:\,\sum_i c_i=0}
\left\{\frac12\sum_i\left(c_i-\frac{A_i}{s_q}\right)^2
+\beta\sum_i c_i\widetilde D_i\right\}.
\end{equation}
The quadratic term controls deviations from the reward coefficients, while the linear term introduces a preference for lower drift. The Lagrangian condition $c_i-A_i/s_q+\beta\widetilde D_i+\lambda=0$ yields $c_i^\star=C_i$, where $\lambda$ is the multiplier for the zero-sum constraint. Because the objective is strictly convex, the ZCPO joint coefficient is the unique optimum of this zero-sum quadratic calibration problem; the full derivation appears in Appendix~\ref{app:coefficient-calibration}. For fixed inputs and under the within-group zero-sum constraint, this result characterizes the optimal trade-off between staying close to the reward coefficients and favoring lower drift.

When $\beta>0$, among responses with equal rewards, those with lower mean KL receive larger joint coefficients, favoring candidates with lower drift at the coefficient level. Adding the same constant to every $\widetilde D_i$ in a group leaves $K_i$ and $C_i$ unchanged, so the coefficients are unaffected by a shared within-group KL baseline (F6). ZCPO uses reference KL for within-group coefficient calibration, and its task-performance advantage over independent KL is evaluated empirically in the tested settings; Appendix~\ref{app:f6-zcpo-calibration} discusses how common-shift invariance relates to reward feedback and candidate coverage.

Within an active group, $\sum_iK_i=\sum_iC_i=0$. At positions where the entire group shares the same context and token, if all gradient gates of the base surrogate are open and the aggregation weights are equal, the direct gradients of both branches cancel together (a strict special case of F2). When the gate associated with the joint coefficient is closed, both branches are zero (F1). The gate $B_q$ directly skips groups with identical rewards (F3). In general, the KL branch of ZCPO can remain nonzero when reward gradients cancel. The joint coefficient affects the clipping branch, so clipped positions need not coincide with those of the independent-KL baseline.

Length normalization converts cumulative KL into a per-token mean, removing the scale factor due to direct accumulation with length (F4 and F5). Conditional KL eliminates the conditional sampling noise of the current token given its prefix; in the construction for F7, $K_i\equiv0$. Variation caused by different sampled prefixes remains. Residual correlations with length and training diagnostics are presented in Section~\ref{sec:exp}.

The optimality and ordering properties of this coefficient calibration do not guarantee that reward increases or KL decreases at every step; when $A_i>0$ and $\beta K_i>A_i/s_q$, we still have $C_i<0$. Appendix~\ref{app:f6-zcpo-calibration} develops an implicit reward curriculum perspective that unifies the roles of KL centering and normalization by the maximum absolute advantage, and discusses their implications for mitigating such reversals. Conditional-KL and k1 coefficients generally induce different raw expected updates; the surrogate update with frozen coefficients is not an unbiased reformulation of the original sequence-KL gradient.

\section{Experiments}
\label{sec:exp}

\paragraph{Experimental setup.} The main mathematical reasoning experiments use Qwen2.5-32B. The vanilla baseline uses DAPO with an independent per-token k3 KL loss, whereas ZCPO integrates our method into DAPO; training and evaluation configurations are provided in Appendix~\ref{app:exp-config}, and the $\beta$ sensitivity analysis is presented in Appendix~\ref{app:beta-sensitivity}. The ablation that replaces conditional KL with k1 in Table~\ref{tab:dapo-zcpo} also uses DAPO and is referred to below as the ``k1 coefficient control.'' Long-context experiments with Qwen3-4B are presented in Appendix~\ref{app:cross-model-task}.

\begin{table}[htbp]
\centering
\caption{Method comparisons, ablations, and F1 controls. Both AIME24 and AIME25 report avg@32 (\%), with means over 3 random seeds. Each ZCPO ablation independently substitutes the definition shown in parentheses; all other training settings match full ZCPO.}
\footnotesize
\label{tab:dapo-zcpo}
\begin{tabular}{lcc}
\toprule
\textbf{Model/method} & \textbf{AIME24 (\%)} & \textbf{AIME25 (\%)} \\
\midrule
DAPO without KL ($\beta=0$) & $47.3$ & $37.1$ \\
DAPO + standard KL, $\beta=0.04$ & $31.2$ & $18.9$ \\
DAPO + standard KL, $\beta=0.001$ & $44.5$ & $34.7$ \\
\midrule
\textbf{ZCPO} & $\mathbf{56.8}$ & $\mathbf{48.5}$ \\
\quad No KL centering ($K_i\leftarrow D_i/T_i$) & $45.6$ & $36.7$ \\
\quad No KL length norm. ($K_i\leftarrow D_i-\frac1G\sum_{j=1}^G D_j$) & $53.9$ & $43.5$ \\
\quad k1 in place of conditional KL ($D_i\leftarrow\sum_{t=1}^{T_i}\Delta_{i,t}$) & $54.5$ & $47.1$ \\
\quad Reward std. norm. ($C_i\leftarrow\hat A_i-\beta K_i$) & $55.4$ & $44.9$ \\
\midrule
RPG & $50.7$ & $42.5$ \\
GVPO & $52.6$ & $41.1$ \\
GOPO & $48.5$ & $39.9$ \\
\midrule
F1-B: Disable KL at clipped tokens ($\ell^{\rm KL}_{i,t}\leftarrow M_{i,t}\ell^{\rm KL}_{i,t}$) & $46.1$ & $39.3$ \\
F1-C: All-token norm control ($\ell^{\rm KL}_{i,t}\leftarrow\operatorname{sg}(\alpha)\ell^{\rm KL}_{i,t}$) & $44.8$ & $37.5$ \\
\bottomrule
\end{tabular}
\par\smallskip
\begin{minipage}{\linewidth}
\footnotesize
Both F1-B and F1-C use $\beta=0.001$, with DAPO + standard KL at the same coefficient in the table as their baseline. They modify only the KL term, retaining the reward branch and aggregation weights.
$M_{i,t}=\operatorname{sg}[m_{\rm base}(r_{i,t},\hat A_i)]$ and
$\alpha=\|g_{\rm KL}^{M}\|_2/\|g_{\rm KL}\|_2$, where $g_{\rm KL}^{M}$ and $g_{\rm KL}$ are the KL-branch parameter gradients after and before gating, respectively, evaluated at the same model state and on the same batch. The ratio is defined only when its denominator is nonzero.
\end{minipage}
\end{table}

\paragraph{Ablation analysis.}
Table~\ref{tab:dapo-zcpo} shows that removing within-group KL centering causes the largest drop in mean accuracy. This ablation both breaks the zero-sum property of the coefficients and retains the shared KL component, supporting the role of centering in the current configuration. Removing KL length normalization, replacing conditional KL with k1, or using standard-deviation reward normalization also yields lower mean accuracy than full ZCPO. These results support the contribution of these components in the current configuration. Because a single component can affect multiple update mechanisms, each ablation evaluates its combined effects.

\begin{figure}[!htb]
\centering
\includegraphics[width=\textwidth]{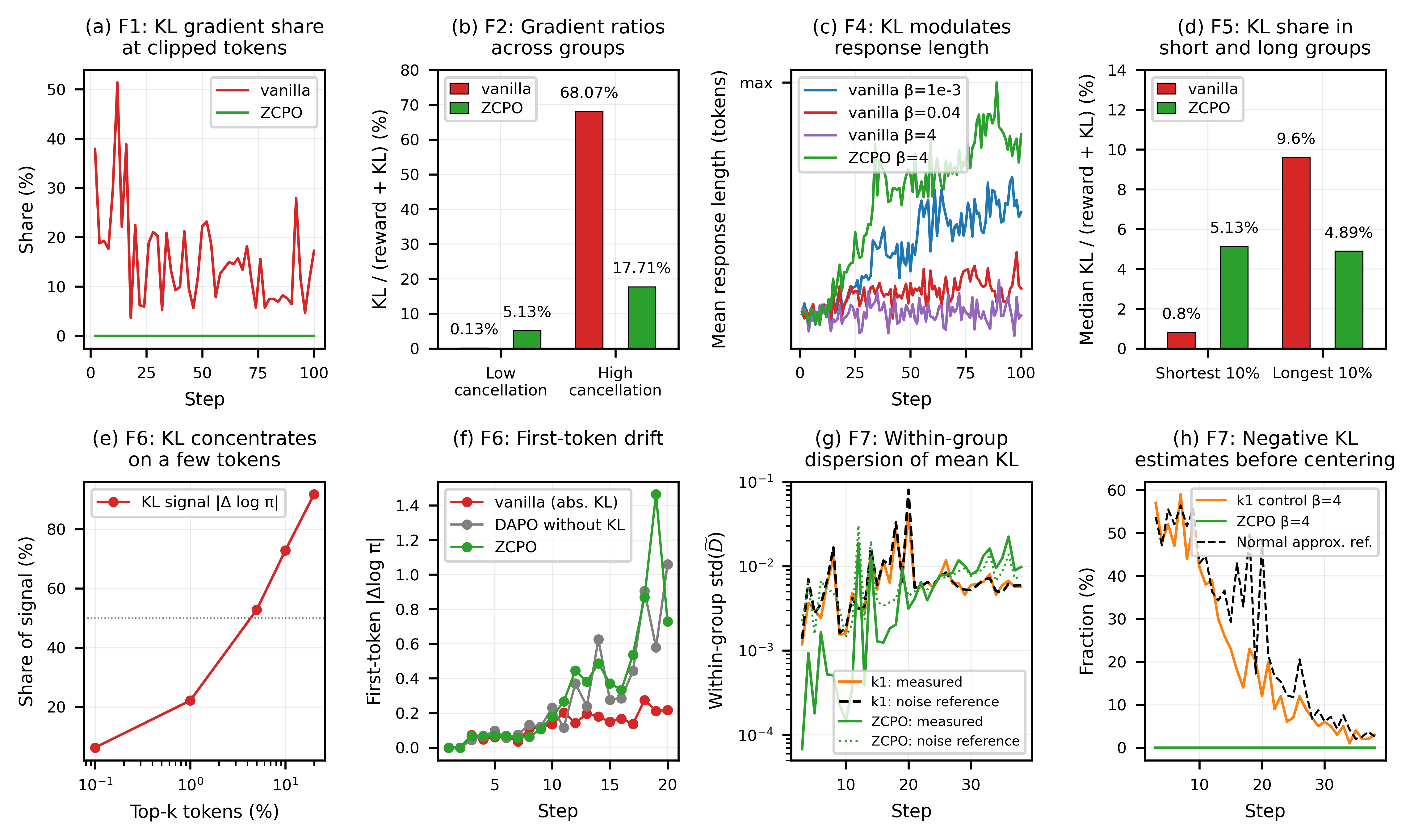}
\caption{KL failure-mode diagnostics. (a) F1: KL gradient share at clipped tokens. (b) F2: KL/total gradient norm ratios for groups in the top/bottom 10\% by reward-gradient cancellation. (c) F4: at step 100, vanilla has mean lengths of 774 and 611 tokens at $\beta=0.04$ and $4$, versus 1695 for ZCPO at $\beta=4$. (d) F5: KL gradient shares in the shortest/longest 10\% of groups. (e,f) F6: KL signal concentration and first-token drift. (g,h) F7: within-group dispersion of mean response KL and negative-estimate fractions, both before centering. Dashed/dotted curves in (g) give small-KL sampling-noise reference scales; the dashed curve in (h) gives the normal approximation for the k1 control. Vanilla denotes DAPO with independent KL loss.}
\label{fig:modes}
\end{figure}

\paragraph{Experimental analysis.} The optimality result in Equation~(\ref{eq:coefficient-calibration}) characterizes zero-sum coefficient calibration for fixed inputs. Table~\ref{tab:dapo-zcpo} shows that ZCPO achieves higher mean task accuracy than DAPO, RPG, GVPO, and GOPO in our setting, and the component ablations further support the key design choices. The changes in gradient contributions, response length, and local drift in Figure~\ref{fig:modes} are consistent with the proposed calibration mechanism. In the figure, vanilla denotes the DAPO + KL-loss baseline described above.

\paragraph{Training performance with larger relative KL gradient contributions.} In the recorded diagnostic runs, the median batch-level KL-to-total gradient norm ratio for ZCPO ($\beta=4$) is approximately $17.5$ times that of vanilla ($\beta=0.001$) (Table~\ref{tab:kl-gradient-share}), while ZCPO maintains strong task performance. The median ratios for vanilla ($\beta=0.04$) and ZCPO ($\beta=4$) are of similar magnitude ($5.8\%$ and $7.0\%$), but the former exhibits substantially restricted task performance and response length. ZCPO also maintains strong task performance at $\beta=20$, where the median ratio reaches $30.1\%$. These results support ZCPO's ability to tolerate larger relative KL gradient contributions in this diagnostic setting.

\paragraph{F1: Residual KL-driven updates after reward clipping.}
As shown in Figure~\ref{fig:modes}(a), vanilla retains KL gradients at tokens whose reward-gradient gates are closed. The ratio of their gradient norm to the KL gradient norm over all tokens reaches approximately 51\%. The KL-gradient overallocation metric for clipped tokens reaches 68--172 times, indicating a substantial residual KL gradient contribution after reward clipping. For ZCPO, the direct gradients of both the reward and KL branches are zero at positions where its own gradient gate is closed, at every recorded step.

Furthermore, the F1 controls in Table~\ref{tab:dapo-zcpo} show that disabling KL at positions clipped by the reward surrogate yields higher mean accuracy than both the original independent-KL baseline and the all-token norm control, supporting the role of selectively disabling KL at these positions in the current setting.

\paragraph{F2: Residual KL after reward-gradient cancellation.}

Let $u^R_{i,t}$ denote the per-token reward-branch parameter gradient, including the reward coefficient, gradient gate, and aggregation weight. For the set $\mathcal I$ of valid tokens included in the statistics, the reward-gradient cancellation rate is defined, when the denominator is nonzero, as
\[
c_R=1-\frac{\left\|\sum_{(i,t)\in\mathcal I}u^R_{i,t}\right\|_2}
{\sum_{(i,t)\in\mathcal I}\left\|u^R_{i,t}\right\|_2}.
\]
The numerator sums the gradient vectors before taking the norm, whereas the denominator sums the norms of individual token gradients. In practice, the denominator is estimated by multiplying the mean gradient norm of tokens sampled uniformly from $\mathcal I$ by $|\mathcal I|$. Under this estimator, the reward-branch gradient cancellation rate reaches 99.6\% during training.

We rank groups by their estimated within-group reward-gradient cancellation rate and sample from the top and bottom 10\% to obtain groups with high and low cancellation, respectively. Across these two selected sets of groups, vanilla's KL gradient share differs by approximately a factor of 524 (Figure~\ref{fig:modes}(b)), indicating a large disparity in the selected groups. The corresponding ratio for ZCPO differs by approximately a factor of 3.45, showing a smaller disparity between the groups.

\textbf{Mechanistic hypothesis:} Reducing disparities in the relative KL contribution across different degrees of reward-gradient cancellation may improve training stability.

\paragraph{F3: Residual KL-driven updates in groups with identical rewards.}
Both vanilla and ZCPO filter out groups with identical rewards before updating in the main mathematical reasoning experiments, so these experiments do not evaluate F3. The long-context experiments in Appendix~\ref{app:cross-model-task} compare full ZCPO against an ablation without gating ($B_q=1$) to evaluate the role of gating groups with identical rewards.

\paragraph{F4: KL modulates language model response length.}
As shown in Figure~\ref{fig:modes}(c), vanilla's mean response length decreases as $\beta$ increases, whereas ZCPO produces longer responses at the same nominal $\beta=4$. This result supports ZCPO's mitigation of the response-length suppression observed in vanilla in our setting. The normalized $K_i$ remains correlated with $T_i$, indicating that length effects are not fully eliminated.

\paragraph{F5: The KL share may increase as more tokens are aggregated.}
Sampling from the shortest and longest 10\% of groups, vanilla's median KL share in the longer groups is 12 times that in the shorter groups. For ZCPO, the medians differ by only 0.24 percentage points, indicating a smaller disparity between the sampled short and long groups (Figure~\ref{fig:modes}(d)).

\paragraph{F6: Potential suppression of beneficial policy changes when KL signals are concentrated.}
The experiments show that a small fraction of tokens accounts for most of the KL signal (Figure~\ref{fig:modes}(e)). Under an independent KL constraint with a fixed reference policy, these positions can continue to experience a local pullback if they reappear in subsequent training and their probabilities still deviate from the reference policy. If this effect conflicts with the direction of reward optimization, it may restrict the corresponding policy changes.

Through within-group centering, ZCPO removes the component of length-normalized KL shared across responses. In our setting, ZCPO exhibits greater drift at the initial token than vanilla, with a magnitude comparable to pure DAPO (Figure~\ref{fig:modes}(f)). At step 20, the covariance contribution of the first 10 tokens to the variance of $K$ is approximately 4\%, indicating that the initial segment does not dominate the variance of the relative KL coefficient at this step. These results support ZCPO's mitigation of drift suppression at initial positions.

\textbf{Mechanistic hypothesis:} If these drifts correspond to beneficial policy changes, mitigating their suppression may help preserve those changes. The analysis is given in Appendix~\ref{app:f6-weak-reward}.

\paragraph{F7: Conditional sampling noise in k1 coefficients.}
Replacing conditional KL $\kappa_{i,t}$ in ZCPO's $D_i=\sum_t\kappa_{i,t}$ with the k1 sampled log-ratio $\Delta_{i,t}$ gives
\[
D_i^{\rm k1}=\sum_{t=1}^{T_i}\Delta_{i,t},\qquad
K_i^{\rm k1}=\frac{D_i^{\rm k1}}{T_i}-\frac1G\sum_{j=1}^G\frac{D_j^{\rm k1}}{T_j}.
\]
The final performance of this control is reported in Table~\ref{tab:dapo-zcpo}, and its training diagnostics appear in Figure~\ref{fig:modes}(g) and (h).
Using the same seed and configuration, we replace $K_i$ with $K_i^{\rm k1}$ as a training control. At steps 3--10, 42\%--59\% of responses in this control receive negative KL estimates before centering, compared with 0 for ZCPO (Figure~\ref{fig:modes}(h)), showing that a single k1 sample can deviate from the nonnegative conditional KL mean. Its within-group dispersion is close to the reference scale under the small-KL approximation (median ratio of 0.95 over steps 3--38; Figure~\ref{fig:modes}(g)), consistent with non-negligible sampling fluctuations.

We further conduct a \textbf{KL-estimator comparison}: fixing the policy, reference policy, and sampled responses recorded in a ZCPO run, we compute $K_i^{\rm k1}$ using the formula above. Its root mean squared error relative to the ZCPO coefficient $K_i$ is approximately 0.00257 and 0.00617 at steps 3 and 8, respectively (Figure~\ref{fig:f7-paired}(a)), indicating that within-group centering does not eliminate the k1 estimation residuals in these batches. By construction, conditional KL in ZCPO eliminates the sampling noise of the current token given its prefix; variation caused by different sampled prefixes remains.

Under the single-token construction for F7, GVPO's centered sequence log-ratio also has nonzero conditional sampling variance~\citep{NEURIPS2025_f21a76d6}. Figure~\ref{fig:f7-paired}(b) shows the GVPO training curves, using the RMS of $D_i^{\rm k1}-D_i$ after within-group centering to measure the magnitude of conditional sampling noise.

\section{Conclusion and Limitations}
\label{sec:conclusion}

We study how reference-policy information can enter group-relative optimization, analyze seven potential failure modes in KL--reward interactions, and propose ZCPO to calibrate within-group zero-sum coefficients using relative drift measured by conditional KL. Mathematical reasoning experiments with Qwen2.5-32B and long-context experiments with Qwen3-4B support this design's effectiveness in our settings; the long-context ablations further evaluate the role of group gating. Our current evaluation covers multiple model scales and tasks; cross-model evidence for model families beyond Qwen is provided in Appendix~\ref{app:glm-cross-model}.

\clearpage
\subsection*{AI Use Statement}

No AI was used.

\subsection*{Ethics Statement}

Not applicable.

\subsection*{Reproducibility Statement}

The supplementary material provides code that can be run directly to reproduce the results in this paper.

\clearpage
\appendix

\section{Experimental Configuration}
\label{app:exp-config}

The base training configuration follows DAPO. Table~\ref{tab:dapo-config} lists the training and evaluation settings and the KL coefficients used in the main experiments. The reward coefficients and KL constructions for vanilla and ZCPO follow the definitions in the main text.
\begin{table}[!ht]
\centering
\caption{Base training and evaluation settings and KL coefficients for the main experiments.}
\label{tab:dapo-config}
\small
\begin{tabular}{@{}p{0.42\textwidth}p{0.54\textwidth}@{}}
\toprule
\textbf{Configuration item} & \textbf{Setting} \\
\midrule
Training framework & verl \\
Training data & DAPO-Math-17K: approximately 17K prompts with integer answers \\
Optimizer & AdamW \\
Learning rate & $1\times10^{-6}$ \\
Learning-rate warmup & Linear warmup over the first 20 rollout steps \\
Learning-rate schedule after warmup & Constant \\
Prompt batch size per rollout & 512 \\
Responses per prompt $G$ & 16 \\
Training mini-batch size & 512 responses \\
Gradient updates per rollout & 16 \\
Lower clipping margin $\varepsilon_{\rm low}$ & 0.2 \\
Upper clipping margin $\varepsilon_{\rm high}$ & 0.28 \\
DAPO base loss aggregation & Token-level mean \\
ZCPO KL coefficient $\beta$ in the main experiments & $4$ \\
Vanilla KL coefficient $\beta$ in the main experiments & $0.001$ \\
Maximum response length without a length penalty & 16,384 tokens \\
Overlong soft-penalty buffer length & 4,096 tokens \\
Maximum generation length & 20,480 tokens \\
Correctness reward & $+1$ for correct answers; $-1$ for incorrect answers \\
Evaluation set & AIME 2024 and AIME 2025 \\
Evaluation samples per problem & 32 \\
Evaluation metric & avg@32 \\
Evaluation temperature & 1.0 \\
Evaluation top-p & 0.7 \\
\bottomrule
\end{tabular}
\end{table}

\paragraph{Checkpoint selection.}
The main experiments are aligned by cumulative GPU compute time, and checkpoints corresponding to the same cumulative GPU compute time are selected for evaluation. Checkpoint selection is independent of the AIME24 and AIME25 evaluation results.

\paragraph{Computation reuse.}
Compared with standard KL, ZCPO uses fused computation to reuse the forward-pass outputs and logits of the current and reference policies, avoiding duplicate full-model forward passes and additional memory for logits. The added arithmetic consists primarily of per-token reductions over the full vocabulary, with essentially negligible additional runtime.
When the conditional distributions of the current and reference policies are close, the exact conditional KL is a second-order quantity in the distribution perturbation and is susceptible to rounding errors in bf16 logits. To ensure numerical accuracy during bf16 training, we recompute the output-layer projection in fp32 and perform the KL reduction over the full vocabulary in fp32; together, this projection recomputation and KL reduction increase total GPU compute time by approximately $0.5\%$ relative to standard KL under the same configuration.

\paragraph{Sampling and filtering.}
In the main mathematical reasoning experiments, we set $\pi_{\rm old}\leftarrow\pi_\theta$ before each rollout. Dynamic sampling filters out response groups in which all answers are correct or all are incorrect, and continues sampling until the prompt batch is filled. Overlong filtering masks the losses of truncated responses.

\paragraph{Overlong soft penalty.}
A length penalty is added to the rule-based correctness reward. Using the response length $T_i$, this penalty is
\[
R_{\rm length}(y_i)=
\begin{cases}
0, & T_i\le16{,}384,\\
\dfrac{16{,}384-T_i}{4{,}096}, & 16{,}384<T_i\le20{,}480,\\
-1, & T_i>20{,}480.
\end{cases}
\]

\clearpage
\section{Sensitivity to \texorpdfstring{$\beta$}{beta}}
\label{app:beta-sensitivity}

The three DAPO configurations in Table~\ref{tab:dapo-zcpo}, with $\beta\in\{0,0.001,0.04\}$, provide a sensitivity analysis of the independent KL loss with respect to $\beta$; $\beta=0$ corresponds to the no-KL control.

Table~\ref{tab:beta-sensitivity} summarizes ZCPO's AIME24 performance at $\beta\in\{0,4,20\}$, following the training and evaluation settings of the main Qwen2.5-32B experiment. At $\beta=0$, normalization by the maximum absolute advantage is retained, giving a control with the KL calibration term disabled.

\begin{table}[!ht]
\centering
\caption{ZCPO sensitivity to $\beta$: AIME24 avg@32 (\%) on Qwen2.5-32B, reporting means over 3 random seeds. The result at $\beta=4$ is reproduced from Table~\ref{tab:dapo-zcpo}.}
\label{tab:beta-sensitivity}
\small
\begin{tabular}{cc}
\toprule
$\boldsymbol{\beta}$ & \textbf{AIME24 (\%)} \\
\midrule
$0$ & $48.1$ \\
$4$ & $56.8$ \\
$20$ & $53.7$ \\
\bottomrule
\end{tabular}
\end{table}

\paragraph{Joint comparison of reward normalization and KL-coefficient calibration.}
Combining Tables~\ref{tab:dapo-zcpo} and~\ref{tab:beta-sensitivity}, Table~\ref{tab:normalization-calibration} organizes four controls by reward normalization and whether KL-coefficient calibration is enabled. The standard-deviation entries correspond to DAPO without KL and the ZCPO ablation using standard-deviation normalization; the maximum-absolute-advantage entries correspond to ZCPO with $\beta=0$ and $\beta=4$.

\begin{table}[!ht]
\centering
\caption{Joint comparison of reward normalization and KL-coefficient calibration. Entries are collected from Tables~\ref{tab:dapo-zcpo} and~\ref{tab:beta-sensitivity}, reporting Qwen2.5-32B AIME24 avg@32 (\%) means over 3 random seeds.}
\label{tab:normalization-calibration}
\small
\begin{tabular}{lcc}
\toprule
\textbf{Reward normalization} & \textbf{KL calibration off} ($\beta=0$) & \textbf{KL calibration on} ($\beta=4$) \\
\midrule
Standard deviation & $47.3$ & $55.4$ \\
Maximum absolute advantage & $48.1$ & $56.8$ \\
\bottomrule
\end{tabular}
\end{table}

Replacing standard-deviation normalization with maximum-absolute-advantage normalization increases mean AIME24 accuracy by $0.8$ and $1.4$ percentage points with KL calibration disabled and enabled, respectively. Conversely, enabling the full KL-coefficient calibration increases mean accuracy by $8.1$ and $8.7$ percentage points under the two normalization schemes. These mean comparisons support gains beyond the change in reward normalization alone in this configuration, with KL-coefficient calibration providing additional gains under both normalization schemes.

\clearpage
\section{Diagnostics of the KL Gradient Share}
\label{app:kl-gradient-share}

\paragraph{Measurement.}
The gradient share is measured on the first mini-batch of each step, when the parameters have not yet been updated in that step, the current policy equals the sampling policy, and every reward gradient gate is open. On the same mini-batch and with the same parameters, two separate full forward--backward passes are run: one with the reward-branch loss only and one with the KL-branch loss only (the k3 term for vanilla, the KL term of the coefficient for ZCPO); the norms of the two parameter gradients and their inner product are read out, and the total gradient norm is composed from them. The share is the ratio of the KL-branch gradient norm to the total gradient norm. Norms are computed over the full parameter vector. These two extra backward passes are used only for recording and do not contribute to the parameter update.

The absolute gradient norms of the two branches show that, within a method, the reward-branch norm stays essentially unchanged across $\beta$ while the KL-branch norm grows approximately linearly with $\beta$, so changes in the share come from the numerator rather than the denominator. ZCPO's reward coefficient is bounded whereas vanilla's standard-deviation-normalized advantage is not, so the absolute norms of the two methods are on different scales and are compared only within a method.

Table~\ref{tab:kl-gradient-share} summarizes the batch-level ratio of the KL gradient norm to the total gradient norm, recorded on the first mini-batch of each step from steps 11--100 in the diagnostic runs, and reports the median over this interval; each configuration has one or two random seeds.

\begin{table}[!ht]
\centering
\caption{Median ratio of the KL gradient norm to the total gradient norm in the diagnostic runs (steps 11--100).}
\label{tab:kl-gradient-share}
\begin{tabular}{lcc}
\toprule
\textbf{Method} & $\boldsymbol{\beta}$ & \textbf{KL/total gradient norm ratio} \\
\midrule
vanilla & 0.001 & 0.4\% \\
vanilla & 0.04 & 5.8\% \\
ZCPO & 4 & 7.0\% \\
ZCPO & 20 & 30.2\% \\
\bottomrule
\end{tabular}
\end{table}

\begin{figure}[!htbp]
\centering
\includegraphics[width=\textwidth]{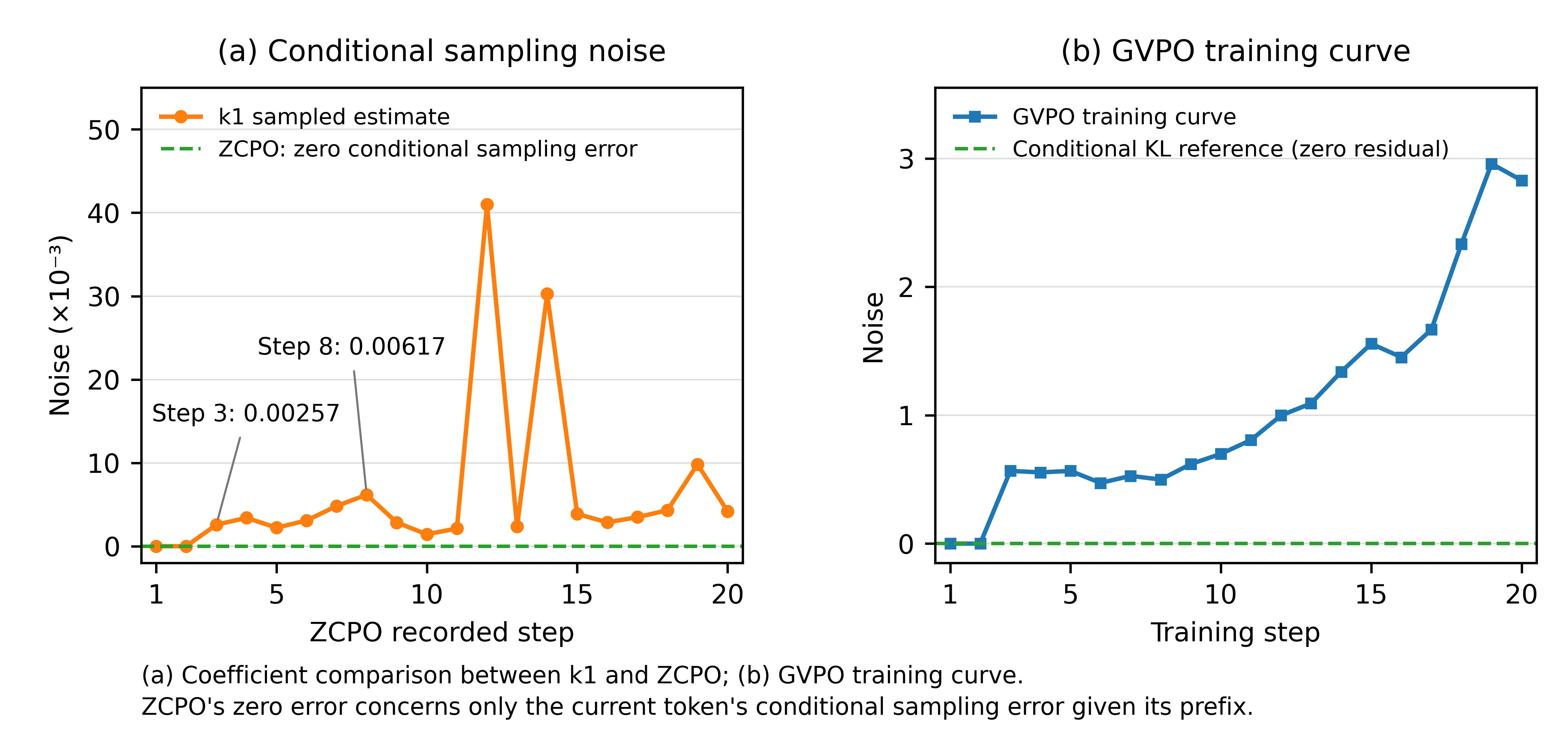}
\caption[]{Conditional sampling-noise comparisons. (a) Using the first mini-batch at each of the first 20 steps of ZCPO, the orange curve shows the residual RMSE of the k1 coefficient relative to ZCPO, and the green curve shows ZCPO's own zero residual. ZCPO's zero conditional sampling error refers only to the local estimation term obtained by replacing sampled $\Delta_{i,t}$ with $\kappa_{i,t}$ for a given prefix; variation in sampled prefixes remains. (b) GVPO training curves. ``Noise'' denotes the sampling residual of the log-ratio relative to its per-prefix conditional mean. Its magnitude is the measured RMS of $D_i^{\rm k1}-D_i=\sum_t(\Delta_{i,t}-\kappa_{i,t})$ after within-group centering, without multiplication by the KL weight. The green dashed line is the zero reference for conditional sampling error.}
\label{fig:f7-paired}
\end{figure}

Figure~\ref{fig:f7-paired} provides supplementary coefficient diagnostics for F7.

\clearpage
\section{Optimality of Zero-Sum Coefficient Calibration}
\label{app:coefficient-calibration}

Fix the samples, rewards, and current policy for an active group, so that $s_q>0$ and both $A_i/s_q$ and $\widetilde D_i$ are treated as constants. The $c_i$ in Eq.~(\ref{eq:coefficient-calibration}) are update coefficients that may be positive or negative. Denote the objective in that equation by $\mathcal Q_q(\boldsymbol c)$ and introduce the Lagrange multiplier $\lambda$ for the zero-sum constraint to obtain
\[
\mathcal H(\boldsymbol c,\lambda)
=\frac12\sum_i\left(c_i-\frac{A_i}{s_q}\right)^2
+\beta\sum_i c_i\widetilde D_i+\lambda\sum_i c_i.
\]
The first-order optimality conditions are
\[
c_i-\frac{A_i}{s_q}+\beta\widetilde D_i+\lambda=0,
\qquad \sum_i c_i=0.
\]
Summing the first equation over $i$ and using $\sum_i A_i=0$ gives
\[
\lambda=-\frac{\beta}{G}\sum_j\widetilde D_j,
\qquad
c_i^\star=\frac{A_i}{s_q}
-\beta\left(\widetilde D_i-\frac1G\sum_j\widetilde D_j\right)=C_i.
\]
This solution satisfies the zero-sum constraint. The Hessian of the objective with respect to $\boldsymbol c$ is the identity matrix, so the objective is strictly convex under this linear constraint, and the solution above is the unique global optimum.

This construction is invariant to a common shift: adding a constant $b$ to every $\widetilde D_i$ changes the objective by $\beta b\sum_i c_i=0$, leaving the optimal coefficients unchanged. For two responses in the same active group, the difference between their optimal coefficients is
\[
C_i-C_j=\frac{R_i-R_j}{s_q}
-\beta\big(\widetilde D_i-\widetilde D_j\big).
\]
Thus, when $\beta>0$ and $R_i=R_j$, $\widetilde D_i<\widetilde D_j$ implies $C_i>C_j$.

This optimality result concerns coefficient calibration with fixed inputs; $\sum_i c_i\widetilde D_i$ is a linear drift score, not a policy KL divergence. The policy parameters $\theta$ are fixed in this problem. During actual training, the optimal coefficients are detached from the gradient computation and inserted into the base surrogate objective in Section~\ref{sec:method}.

\clearpage
\section{Cross-Model and Cross-Task Experiments}
\label{app:cross-model-task}

Starting from Qwen3-4B-Thinking-2507, we conduct long-context training on a fixed subset of 8,000 examples from the GoLongRL dataset~\citep{lv2026golongrlcapabilityorientedlongcontext}. We compare GRPO with standard KL, GRPO without KL, GRPO-ZCPO, and its group-gating ablation on six benchmarks. The standard KL baseline uses a coefficient of $0.001$, and the $\beta$ values for full ZCPO and the ablation without gating are both $4$.

Full ZCPO retains the $B_q$ gate defined in the main text. The ablation without gating sets $B_q=1$ and sets the reward coefficient to zero when $s_q=0$:
\[
a_i^{\rm abl}=\begin{cases}
A_i/s_q, & s_q>0,\\
0, & s_q=0,
\end{cases}
\qquad C_i^{\rm abl}=a_i^{\rm abl}-\beta K_i.
\]

\begin{table}[!ht]
\centering
\begin{threeparttable}
\caption{Long-context evaluation results and group-gating ablation for Qwen3-4B. All methods trained in this work use the same data subset. $B_q=1$ denotes the ablation that removes gating for groups with identical rewards, with all other ZCPO settings unchanged.}
\label{tab:main_results}
\footnotesize
\setlength{\tabcolsep}{4pt}
\renewcommand{\arraystretch}{1.1}
\begin{tabular}{lcccccc}
\toprule
\textbf{Model} & \textbf{DocMath} & \textbf{LBV2}
  & \textbf{Frames} & \textbf{MRCR} & \textbf{CorpusQA} & \textbf{LBV1-QA} \\
\midrule
Qwen3-4B-Thinking-2507
  & 60.7 & 39.4 & 64.9 & 37.5 & 49.2 & 64.2 \\
+ \textit{(GRPO + standard KL)}~
  & 61.6 & 41.7 & 64.8 & 49.3 & 53.8 & 64.3 \\
+ \textit{(GRPO without KL)}~
  & 62.3 & 45.2 & 66.4 & 65.7 & 61.9 & 65.6 \\
+ \textit{(GRPO-ZCPO)}
  & \textbf{63.8} & \textbf{48.2} & \textbf{68.2} & 66.0 & \textbf{69.6} & \textbf{66.4} \\
+ \textit{(GRPO-ZCPO, $B_q=1$)}
  & 63.3 & 48.0 & 67.6 & \textbf{66.3} & 68.5 & 66.1 \\
\bottomrule
\end{tabular}\end{threeparttable}
\end{table}

\clearpage
\subsection{Training Hyperparameters for Long-Context Experiments}
\label{app:golongrl-hyperparameters}

Table~\ref{tab:hyperparams} summarizes the training hyperparameters for the long-context experiments. The KL and group-filtering settings for each method are listed in Table~\ref{tab:golongrl-script-config}.

\paragraph{Evaluation tasks.}
The long-context evaluation includes DocMath, LongBench-V2, Frames, MRCR, CorpusQA, and LongBench-v1 QA (LBV1-QA). The last benchmark comprises five subsets: 2WikiMultihopQA, HotpotQA, MuSiQue, NarrativeQA, and Qasper.

\begin{table}[!ht]
\centering
\small
\caption{Training hyperparameters for the long-context experiments.}
\label{tab:hyperparams}
\setlength{\tabcolsep}{10pt}
\renewcommand{\arraystretch}{1.15}
\begin{tabularx}{\linewidth}{@{}Xr@{}}
\toprule
\textbf{Hyperparameter} & \textbf{Value} \\
\midrule
\multicolumn{2}{@{}l}{\textit{Data}} \\
\quad Max prompt length & 160K \\
\quad Max response length & 16K \\
\quad Responses per prompt & 16 \\
\midrule
\multicolumn{2}{@{}l}{\textit{Optimization}} \\
\quad Learning rate & 2e-6 \\
\quad LR warmup steps & 5 \\
\quad Weight decay & 0.1 \\
\quad Gradient clipping & 1.0 \\
\quad PPO epochs & 1 \\
\quad Total epochs & 10 \\
\quad Train prompt batch size & 128 \\
\quad Loss aggregation & token-mean \\
\midrule
\multicolumn{2}{@{}l}{\textit{Clipping}} \\
\quad Clip ratio (low) & 0.2 \\
\quad Clip ratio (high) & 0.28 \\
\quad Clip ratio $c$ & 3.0 \\
\midrule
\multicolumn{2}{@{}l}{\textit{Importance Sampling}} \\
\quad IS level & token \\
\quad IS clipping mode & clip \\
\quad IS threshold (upper) & 5.0 \\
\quad IS threshold (lower) & 0.5 \\
\quad IS veto threshold & 1e-4 \\
\bottomrule
\end{tabularx}
\end{table}

\clearpage
\subsection{Additional Configuration for Long-Context Experiments}
\label{app:golongrl-script-config}

Table~\ref{tab:golongrl-script-config} provides the model, KL, sampling, parallelism, and checkpoint settings.

\begingroup
\small
\setlength{\tabcolsep}{4pt}
\renewcommand{\arraystretch}{1.05}
\begin{longtable}{@{}p{0.43\linewidth}p{0.53\linewidth}@{}}
\caption{Additional configuration for the Qwen3-4B long-context experiments. KL and group-gating settings that differ across methods are listed separately.}
\label{tab:golongrl-script-config}\\
\toprule
\textbf{Parameter} & \textbf{Setting} \\
\midrule
\endfirsthead
\multicolumn{2}{c}{Table~\ref{tab:golongrl-script-config} (continued)}\\
\toprule
\textbf{Parameter} & \textbf{Setting} \\
\midrule
\endhead
\bottomrule
\endfoot
\multicolumn{2}{@{}l}{\textbf{Model, Data, and Method}}\\
Starting model & Qwen3-4B-Thinking-2507 \\
Training data & A fixed subset of 8,000 GoLongRL examples; the same subset for all methods \\
Training framework and entry point & verl; \texttt{recipe.dapo.main\_dapo} \\
Trainer configuration & \texttt{dapo\_megatron\_trainer} \\
Update coefficient & GRPO uses group-relative advantages; ZCPO uses the joint coefficient $C_i$ defined in the main text \\
Additional difficulty weighting & Disabled \\
\midrule
\multicolumn{2}{@{}l}{\textbf{Objective, KL, and Filtering}}\\
Policy loss mode & \texttt{vanilla} \\
KL added to rewards & \texttt{use\_kl\_in\_reward=False} \\
Reward KL coefficient & \texttt{kl\_coef=0.0} \\
Separate KL loss & Enabled only for the GRPO baseline with standard KL; disabled for GRPO without KL and ZCPO \\
Standard KL baseline coefficient & $0.001$ \\
ZCPO $\beta$ & $4$ (for both the full method and the ablation without gating) \\
Entropy regularization coefficient & \texttt{entropy\_coeff=0} \\
Prefilter groups with identical rewards & Disabled (\texttt{filter\_groups.enable=False}) \\
Internal ZCPO group gating & The full method retains $B_q$; the ablation without gating sets $B_q=1$ \\
Reward coefficient for groups with identical rewards in the ablation without gating & Set to zero when $s_q=0$ \\
Rollout importance sampling correction & Enabled; see Table~\ref{tab:hyperparams} for other thresholds \\
Reward manager & \texttt{dapo} \\
Soft overlength penalty & Disabled (\texttt{overlong\_buffer.enable=False}) \\
Overlength buffer / penalty coefficient & 4,096 tokens / 1.0; inactive because the option is disabled \\
\midrule
\multicolumn{2}{@{}l}{\textbf{Generation, Batch Sizes, and Lengths}}\\
Training sampling temperature / top-p & 1.0 / 1.0 \\
Training sampling top-k & $-1$ (top-k truncation disabled) \\
Generation prompt batch size & 512 \\
Training prompt batch / PPO mini-batch & 128 / 128 \\
PPO micro-batch per GPU & 1 \\
Dynamic batch size & Disabled \\
Maximum prompt / response length & 163,840 / 16,384 tokens \\
Maximum model length for rollouts & 180,224 tokens \\
Maximum assistant turns per rollout & 1 \\
Overlength prompt filtering & \texttt{filter\_overlong\_prompts=True} \\
Data truncation direction & \texttt{truncation=left} \\
Data order / prompt field & \texttt{shuffle=True}; \texttt{prompt} \\
Filtering cache & \texttt{use\_filtered\_cache=True}; \texttt{rebuild\_filtered\_cache=False} \\
\midrule
\multicolumn{2}{@{}l}{\textbf{Parallelism and Runtime Configuration}}\\
GPU nodes / GPUs per node & 16 / 8 (128 GPUs in total) \\
Training / inference backend & Megatron (mbridge enabled) / SGLang \\
Rollout service mode & \texttt{async} \\
Training tensor parallelism TP & 8 \\
Training expert parallelism EP & 1 \\
Context / pipeline parallelism CP / PP & 8 / 1 \\
Expert tensor parallelism ETP & 1 \\
Inference tensor parallelism & 8 \\
Actor/ref parameter, gradient, optimizer offload & All \texttt{True} \\
Attention backend / padding removal & \texttt{flash} / \texttt{True} \\
Actor activation recomputation & \texttt{uniform}, \texttt{full}, 1 layer per group \\
PPO / log-prob token budget per GPU & 22,528 / 22,528; dynamic batch size disabled \\
Log-prob micro-batch / dynamic batching & 1 / \texttt{False} \\
Rollout GPU memory utilization & 0.6 \\
Rollout workers / buffer capacity & 16 / 32 \\
Maximum concurrent requests / CUDA graph batch & 512 / 512 \\
Chunked prefill / rollout log-prob & Both enabled \\
Weight update bucket size & 2,048 MB \\
Actor/ref Torch compile & Both disabled \\
\midrule
\multicolumn{2}{@{}l}{\textbf{Checkpoints and In-Script Validation}}\\
Checkpoint saving interval & Every 5 training steps \\
Maximum actor checkpoints retained & 60 \\
Resume mode / total epochs & \texttt{auto} / 10 \\
Validation before training & \texttt{val\_before\_train=False} \\
Periodic validation frequency & \texttt{test\_freq=-1} (periodic validation disabled) \\
Validation sampling temperature / top-p & 0.6 / 0.95 \\
Validation top-k / do-sample / samples & $-1$ / \texttt{True} / 1 \\
Validation generations logged & 15 \\
Logging backends & \texttt{console}, \texttt{tensorboard} \\
\end{longtable}
\endgroup

\clearpage
\section{Cross-Model Generalization: GLM-4-9B-0414 Math Experiments}
\label{app:glm-cross-model}

To examine ZCPO on models outside the Qwen family, we use GLM-4-9B-0414~\citep{glm2024chatglmfamilylarge} as the training start for math RLVR, comparing DAPO without KL against ZCPO built on DAPO (i.e., our method). Apart from the training start, the training and evaluation settings follow the main experiments (Appendix~\ref{app:exp-config}), with $\beta=4$ for ZCPO; the metric is AIME24 avg@32, reporting means.

\begin{table}[!ht]
\centering
\caption{Cross-model math results on GLM-4-9B-0414 (AIME24 avg@32, \%).}
\label{tab:glm-cross-model}
\small
\begin{tabular}{lc}
\toprule
\textbf{Model/method} & \textbf{AIME24 avg@32 (\%)} \\
\midrule
GLM-4-9B-0414 + DAPO without KL & $24.5$ \\
GLM-4-9B-0414 + ZCPO & $\mathbf{36.2}$ \\
\bottomrule
\end{tabular}
\end{table}

On this non-Qwen base, adding ZCPO improves the AIME24 mean accuracy over DAPO without KL, with a trend consistent with the main experiments, providing preliminary support for ZCPO's cross-family generalization.

\clearpage
\section{Independent KL Suppression of Weak Reward Signals}
\label{app:f6-weak-reward}

The following construction fixes samples, rewards, the old policy, and the reference policy, and demonstrates both reversal directions and the trade-off in global $\beta$ adjustment without clipping.

\paragraph{Reversal of a weak continuous-reward signal.}
Let $G=16$ and let the group rewards consist of seven pairs $(1,-1)$ and one pair $(\tau,-\tau)$, where $0<\tau<1$. The group mean is zero. With sample-standard-deviation normalization,
\[
\sigma_q=\sqrt{\frac{14+2\tau^2}{15}},\qquad
\hat A_{\tau}=\frac{\tau}{\sigma_q+\varepsilon_{\rm adv}}
\longrightarrow0\quad(\tau\longrightarrow0).
\]
For a token in the response with reward $\tau$, the independent $k_3$ KL loss in ordinary GRPO is $\ell=-\hat A_{\tau}r+\beta(e^{-\Delta}+\Delta-1)$. Writing $g=\nabla_\theta\log\pi_\theta(y\mid h)$, its direct gradient-descent contribution at $r=1$ is
\begin{equation}
\label{eq:f6-weak-gradient}
-\nabla_\theta\ell=
\bigl[\hat A_{\tau}-\beta(1-e^{-\Delta})\bigr]g.
\end{equation}
Set the current and old probabilities to $0.4$ and the reference probability to $0.2$, giving $\Delta=\log2$. For any $\beta>0$, a sufficiently small $\tau$ gives $\hat A_\tau<\beta/2$, making the coefficient in Equation~(\ref{eq:f6-weak-gradient}) negative. With $\tau=10^{-3}$, $\varepsilon_{\rm adv}=10^{-6}$, and $\beta=0.04$, we obtain $\hat A_\tau\approx0.0010351$ and a combined coefficient of approximately $-0.0189649$.

\paragraph{The corresponding negative-to-positive reversal.}
The response with reward $-\tau$ in the same group has $\hat A_{-\tau}=-\hat A_\tau$. Set its token's current and old probabilities to $0.6/15$ and the reference probability to $0.8/15$. Then $r=1$, $\Delta=\log(0.75)$, and its direct update coefficient is $h_{-\tau}=-\hat A_\tau+\beta/3$. For any $\beta>0$, a sufficiently small $\tau$ makes this positive; the values above give $h_{-\tau}\approx+0.0122982$. Independent KL can therefore turn a negative reward contribution positive. Both examples are unclipped mathematical constructions concerning direct token contributions, not overall training performance.

\paragraph{Global $\beta$ adjustment under uniform and concentrated contributions.}
For the unclipped surrogate, consider only positions where KL opposes the reward direction and $g_{i,t}\ne0$, namely $\hat A_i\Delta_{i,t}>0$. Define the local KL gradient coefficient before multiplication by $\beta$ and its strength relative to the reward coefficient:
\[
v_{i,t}=|1-e^{-\Delta_{i,t}}|,\qquad
\lambda_{i,t}=\frac{v_{i,t}}{|\hat A_i|r_{i,t}}.
\]
The two terms for a token share its score gradient and aggregation weight, so their contribution-norm ratio is $\beta\lambda_{i,t}$. Avoiding local direction reversal at all such positions requires
\begin{equation}
\label{eq:f6-global-beta}
\beta\le\frac{1}{\max_{i,t}\lambda_{i,t}},
\end{equation}
where the maximum is restricted to the conflicting positions above. If the $\lambda_{i,t}$ values are approximately uniform, lowering the global $\beta$ mitigates suppression comparably across positions. If relative KL contributions concentrate at a few conflicting positions, lowering $\beta$ to protect them also weakens regularization elsewhere: $\beta$ uniformly scales $\beta\lambda_{i,t}$ without changing the relative disparity across positions.

For two conflicting positions with the same $\hat A_\tau$ and $r=1$, let $v_{\rm high}=0.5$ and $v_{\rm low}=0.005$, giving $\lambda_{\rm high}=100\lambda_{\rm low}$. Protecting the former requires $\beta\le2\hat A_\tau\approx0.0020702$, limiting the other position's KL-to-reward ratio to $0.01$; it cannot simultaneously remain at least $0.1$. This is a trade-off in global tuning. Here concentration refers to relative gradient contributions that oppose rewards, which cannot be inferred from scalar KL concentration alone.

\clearpage
\subsection{How ZCPO Mitigates This Failure Mode}
\label{app:f6-zcpo-calibration}

ZCPO combines relative drift with normalization by the maximum absolute advantage to remove the direct coefficient penalty from shared drift and turn reference information into a signal for comparing candidates within a group. Write $a_i=A_i/s_q$ and decompose the mean conditional KL as $\widetilde D_i=b_q+\delta_i$, where $b_q$ is common to the group. Then
\[
K_i=\widetilde D_i-\frac1G\sum_j\widetilde D_j
=\delta_i-\bar\delta,\qquad
C_i=a_i-\beta(\delta_i-\bar\delta).
\]
The shared drift $b_q$ no longer enters the coefficient penalty, while informative relative differences remain. For equal rewards, $C_i-C_j=-\beta(\widetilde D_i-\widetilde D_j)$: responses with lower drift receive larger coefficients, turning reference information into a relative preference.

\paragraph{Common-shift invariance and reward feedback.}
An independent KL term against a fixed reference penalizes policy deviation without evaluating its task benefit, so a beneficial change that increases KL also incurs a regularization cost. ZCPO removes the direct coefficient penalty from the shared component $b_q$; this algebraic property does not distinguish whether a shared change is beneficial. Both approaches retain reward-driven updates. Because different prompts share policy parameters, the absence of reward comparisons in the current group does not imply an absence of reward feedback throughout training: when the reward captures a behavior's effects and other prompts or later rollouts provide informative alternatives, corresponding reward comparisons may still arise. This describes a possible source of learning signals, not an established effect of correction across groups. If a change leaves the reward unchanged across the relevant training distribution and its effect on KL is entirely shared within each group, the reward provides no direct signal to distinguish that change, and the common KL component removed by centering provides no direct pull toward the reference policy.

\paragraph{Group size and the probability of reward comparisons.}
Fix the prompt, policy, and sampling configuration for each response, and consider binary rewards with probability $p\in(0,1)$ of sampling a high-reward response. Under $G$ independent and identically distributed draws, the probability that a group contains both high- and low-reward responses and its increment are
\[
\begin{gathered}
P_{\rm mixed}(G)=1-p^G-(1-p)^G,\\
P_{\rm mixed}(G+1)-P_{\rm mixed}(G)
=p^G(1-p)+(1-p)^G p>0.
\end{gathered}
\]
Thus, under this fixed distribution, increasing $G$ raises the probability of obtaining reward comparisons within a single group. This is a statement about sampling coverage, not training performance after changing $G$; the shared KL component $b_q$ still cancels from $K_i$ for any $G$.

\paragraph{Restoring the weak reward direction in the same example.}
Instantiate the preceding construction with 16 single-token responses to one prompt and a vocabulary of 16 tokens. For the token $y_\tau$ with weak positive reward, set the current and old probabilities to $0.4$ and the reference probability to $0.2$. Assign each remaining token probability $0.6/15$ under the current and old policies and $0.8/15$ under the reference. Consider a sampled group containing each token once, with the preceding seven reward pairs $(1,-1)$ and one pair $(\tau,-\tau)$.

All responses share the same prefix, so their full-vocabulary conditional KL is identical:
\[
\begin{gathered}
\widetilde D_i=d_0=0.4\log2+0.6\log(0.75)\approx0.10465,\\
K_i=0,\qquad s_q=1,\qquad C_\tau=\tau>0,\qquad C_{-\tau}=-\tau<0.
\end{gathered}
\]
For the same policies, sampled group, and $\beta=0.04$, setting $\tau=10^{-3}$ gives local coefficients of approximately $-0.0189649$ and $+0.0122982$ for the responses with weak positive and negative advantages under ordinary k3, versus $+0.001$ and $-0.001$ under ZCPO. Removing shared drift restores both reward directions without reducing the global $\beta$ to accommodate the common component.

\paragraph{Fixing the strongest reward signal to stabilize relative calibration.}
Normalization by the maximum absolute advantage fixes the strongest reward signal in every active group, preventing its peak from varying with the within-group reward distribution:
\[
\|\boldsymbol a\|_\infty=1,\qquad
\rho_{\infty,q}
=\frac{\beta\|\boldsymbol K\|_\infty}{\|\boldsymbol a\|_\infty}
=\beta\|\boldsymbol K\|_\infty.
\]
Here $\rho_{\infty,q}$ is the peak coefficient ratio between the KL correction and the reward term. Variation in the reward peak is therefore removed as a source of fluctuation in relative KL calibration strength, leaving that strength determined by $\beta$ and the actual within-group drift differences. Standard-deviation normalization uses a second-order scale, whereas normalization by the maximum absolute advantage fixes the peak scale.

\paragraph{An implicit reward curriculum: from dominant signals to residual differences.}
ZCPO's within-group calibration can be viewed as an implicit reward curriculum. Consider $R_i=\alpha x_i+\eta z_i$ with $\alpha\gg\eta>0$. When the dominant factor becomes shared, $x_i=x_\star$, its reward contribution cancels. If $z_i$ still varies, we obtain
\[
C_i=
\frac{z_i-\bar z}{\max_j|z_j-\bar z|}
-\beta(\delta_i-\bar\delta).
\]
The original scale $\eta$ cancels, restoring the strongest remaining reward signal to unit magnitude, while relative KL removes the shared drift $b_q$. Factors that become common leave the within-group comparison, and residual differences gain renewed relative weight without an additional curriculum scheduler. This coefficient-level reweighting offers a perspective on how the focus of updates can shift from dominant signals to previously masked differences.

These properties concern calibration coefficients; the actual KL gradient share also depends on relative drift, clipping, and gradient aggregation.

\clearpage
\section{Mini-batch Updates with Frozen Coefficients}
\label{app:minibatch-updates}

In the main experiments, full ZCPO computes coefficients once per rollout. Let $\theta_k$ be the sampling-policy snapshot, $\pi_{\rm old}^{(k)}=\pi_{\theta_k}$, with a fixed reference policy. We write the coefficient construction for one active prompt group.

\paragraph{Computing coefficients at the sampling-policy snapshot.}
Before parameter updates, we reuse the old-policy and reference forward passes to compute
\[
\begin{gathered}
\kappa_{i,t}^{(k)}=\mathrm{KL}\!\left(\pi_{\theta_k}(\cdot\mid h_{i,t})\|\pi_{\rm ref}(\cdot\mid h_{i,t})\right),\qquad
\widetilde D_i^{(k)}=\frac1{T_i}\sum_t\kappa_{i,t}^{(k)},\\
K_i^{(k)}=\widetilde D_i^{(k)}-\frac1G\sum_j\widetilde D_j^{(k)},\qquad
C_i^{(k)}=\frac{A_i}{s_q}-\beta K_i^{(k)}.
\end{gathered}
\]
Rewards, groups, and these coefficients stay fixed through the rollout's mini-batch updates and are refreshed for the next rollout. The $B_q$ gate skips groups with identical rewards.

\paragraph{Mini-batch updates with fixed coefficients.}
Each rollout's training batch contains $512\times16=8192$ responses. One PPO epoch uses 16 mini-batches of 512 responses, preserving prompt groups. For active responses $\mathcal B_{k,b}$ in mini-batch $b$, the surrogate is
\[
\begin{gathered}
r_{i,t}^{(k)}(\theta)=\frac{\pi_\theta(y_{i,t}\mid h_{i,t})}{\pi_{\theta_k}(y_{i,t}\mid h_{i,t})},\\
\mathcal L_{k,b}(\theta)=-\sum_{i\in\mathcal B_{k,b}}\sum_t
w_{i,t}^{(k,b)}S_{\rm base}\!\left(r_{i,t}^{(k)}(\theta);\operatorname{sg}(C_i^{(k)})\right).
\end{gathered}
\]
Weights $w_{i,t}^{(k,b)}$ follow the base aggregation rule. Updates change $\theta$ and the probability ratios, keeping $\theta_k$ and $C_i^{(k)}$ fixed. The clipping gate is evaluated from the current ratio and the fixed joint coefficient, so its state can change during optimization.

\paragraph{Roles of KL computation and the outer probability ratio.}
Conditional KL is evaluated by full-vocabulary summation at the snapshot; the outer ratio and clipping govern updates to $\pi_\theta$. For the paired log-ratio $\Delta^{(k)}(y,h)=\log[\pi_{\theta_k}(y\mid h)/\pi_{\rm ref}(y\mid h)]$, a token drawn from $\pi_{\theta_k}(\cdot\mid h)$ satisfies
\[
\mathbb E_{y\sim\pi_{\theta_k}(\cdot\mid h)}\!\left[\Delta^{(k)}(y,h)\right]
=\mathrm{KL}\!\left(\pi_{\theta_k}(\cdot\mid h)\|\pi_{\rm ref}(\cdot\mid h)\right).
\]
This is a conditional estimation identity at the sampling snapshot. Later updates do not change the frozen quantities' target, so no extra importance correction is needed for subsequent changes in $\theta$. Equation~(\ref{eq:coefficient-calibration}) calibrates coefficients; $\mathcal L_{k,b}$ defines the policy update.

\section{Evidence Scope and Discussion}
\label{app:evidence-scope}

We study the use of reference-policy information at three levels: update structure, coefficient construction, and task performance. The theoretical analysis characterizes the interaction between KL and reward updates under specified conditions and gives the optimal zero-sum coefficient calibration for fixed inputs. The F1 mechanism controls and F3 group-gating ablation evaluate the roles of KL updates at clipped positions and gating groups with identical rewards, respectively, in their corresponding settings. Other gradient, length, and drift diagnostics characterize training behavior and motivate testable mechanistic hypotheses about its effects on optimization. Mathematical reasoning, long-context, and cross-model experiments jointly support ZCPO's task performance in the evaluated configurations. These cross-setting results assess the effectiveness of the full method, while the scope and contribution of individual mechanisms to task gains are discussed separately through analyses and controls under their respective conditions.

\end{document}